\documentclass[letterpaper, 10 pt, conference]{ieeeconf}  

\IEEEoverridecommandlockouts                              

\usepackage{graphicx}

\usepackage{multirow}
\usepackage{biblatex}
\title{\LARGE \bf
ART-SLAM: Accurate Real-Time 6DoF LiDAR SLAM}

\author{Matteo Frosi$^{1*}$, Matteo Matteucci$^{2}$
\thanks{*Corresponding author}
\thanks{$^{1}$Matteo Frosi is a Ph.D. student at the the Dipartimento di Elettronica Informazione e Bioingegneria of Politecnico di Milano, Milan, Italy,
        {\tt\small matteo.frosi@polimi.it}}%
\thanks{$^{2}$Matteo Matteucci is Associate Professor at the Dipartimento di Elettronica Informazione e Bioingegneria of Politecnico di Milano,
        Politecnico di Milano, Italy,
        {\tt\small matteo.matteucci@polimi.it}}%
}

\begin{document}

\maketitle
\thispagestyle{empty}
\pagestyle{empty}

\begin{abstract}

Real-time six degree-of-freedom pose estimation with ground vehicles represents a relevant and well studied topic in robotics, due to its many applications, such as autonomous driving and 3D mapping. Although some systems exist already, they are either not accurate or they struggle in real-time setting. In this paper, we propose a fast, accurate and modular LiDAR SLAM system for both batch and online estimation. We first apply downsampling and outlier removal, to filter out noise and reduce the size of the input point clouds. Filtered clouds are then used for pose tracking and floor detection, to ground-optimize the estimated trajectory. The availability of a pre-tracker, working in parallel with the filtering process, allows to obtain pre-computed odometries, to be used as aids when performing tracking. Efficient loop closure and pose optimization, achieved through a g2o pose graph, are the last steps of the proposed SLAM pipeline. We compare the performance of our system with state-of-the-art point cloud based methods, LOAM, LeGO-LOAM, A-LOAM, LeGO-LOAM-BOR and HDL, and show that the proposed system achieves equal or better accuracy and can easily handle even cases without loops. The comparison is done evaluating the estimated trajectory displacement using the KITTI and RADIATE datasets.

\end{abstract}

\vspace{5px}

\section{INTRODUCTION} \label{introduction}

Trajectory estimation and map building represent core aspects of many applications in robotics, such as autonomous driving.
A great amount of simultaneous localization and mapping (SLAM) systems with 6 degree-of-freedom (6-DoF) have been proposed in literature in the last decades, with the goal of estimating accurate trajectories with real-time performances. These methods can be grouped in two main categories, vision-based and point cloud-based systems, depending on the main sensor used (camera or laser rangefinder, respectively).

Point cloud-based systems, can capture and represent the environment with a high level of details, due to the density of the clouds, and they are not afflicted by the issues of vision-based methods, such as illumination and viewpoint changes. Moreover, tracking performed with point clouds is more accurate and stable than its visual counterpart, and it is generally preferred when data is available. However, achieving real-time performance, while keeping high accuracy, still remains an open quest.

Indeed, the most critical aspect which minders real-time point cloud SLAM is the alignment of LiDAR scans. During the last decades, many algorithms have been created to find the relative motion between two point clouds, operation also known as scan matching. The most used and known method to perform scan matching is Iterative Closest Point (ICP) \cite{besl1992method} and its many variants. The idea behind these algorithms is to align two point clouds iteratively, until convergence or a stopping criterion is satisfied. Although ICP suffers from high computational cost, Generalized ICP \cite{segal2009generalized} and more recent parallel versions (e.g. VGICP \cite{koide2020voxelized}) are relatively faster, and can be used as alternatives. 

To overcome the computational shortcomings of full point clouds scan matching, different feature-based approaches have been proposed. Feature-based matching methods work similarly to standard scan matching, but require less resources. They can achieve so by extracting 3D features from the clouds, such as edges, planes or clusters, and then match them. A low-drift feature-based and real-time LiDAR odometry and mapping (LOAM) method is proposed in \cite{zhang2014loam}. LOAM performs 3D point feature to edge, and plane, scan matching to find correspondences between point clouds. The performance of LOAM deteriorates when resources are limited and no loop closure is performed, leading to large estimation errors on long trajectories, as we show in Section \ref{experiments}. Improving LOAM, the same authors proposed LeGO-LOAM \cite{shan2018lego}, a lightweight, real-time pose estimation and mapping system, composed by five modules: segmentation, feature extraction, LiDAR odometry, LiDAR mapping and transform integration. Speedup is achieved by filtering the input point clouds through image-based segmentation, performed on the 2D range projection of each scan. More recent variants of LeGO-LOAM have been proposed, optimizing it, namely A-LOAM and LeGO-LOAM-BOR.

Feature-based systems are, in general, less accurate than methods which perform scan matching on whole clouds. For this reason, loop closure and trajectory optimization are mandatory steps in their pipeline. Nevertheless, these tasks can easily become computationally demanding as the size of a trajectory increases. To overcome this problem, graph SLAM systems have been proposed, such as \cite{mendes2016icp} and \cite{pierzchala2018mapping}, where the trajectory of the robot, estimated via scan matching, is modeled as a graph. There are multiple advantage of this approach, as described by Grisetti et. al. in \cite{grisetti2010tutorial}, such as the ability to introduce relationships between sensor data and/or observations from the environment, or a great availability of frameworks for efficient graph optimization, which translates in the optimization of the corresponding trajectory.

A recent point cloud-based system which relies on a graph structure is HDL \cite{koide2018portable}, which consists of four steps. First, laser scans are pre-processed and filtered to reduce their size. Then, the filtered clouds are used to simultaneously perform tracking and to possibly detect the ground plane. Poses estimated through tracking, and floor coefficients extracted from the point clouds, are used to build a graph of the trajectory, i.e., a pose graph , which is later optimized. The system achieves superior performances, but it is slow, especially when dealing with large point clouds derived from outdoor environments.

All the algorithms available in literature, being based on feature matching or full scan matching, either achieve high accuracy at the cost of computational time, or sacrifice the quality of the trajectory to obtain real-time performance. Moreover, these systems are monolithic and difficult to modify and adapt, and are usually bound to some existing framework (e.g., ROS \cite{quigley2009ros}), often hindering portability on different operative systems (e.g. Windows or macOS).

For these reasons, in this paper we propose a new system, \textit{ART-SLAM}, to perform point cloud-based graph SLAM, inspired by HDL, with multiple contributions. ART-SLAM is able to achieve real-time performance, retaining high accuracy, even in scenarios without loops. The proposed system is also able to efficiently detect and close loops in the trajectory, using a three-phased algorithm. ART-SLAM presents a high degree of modularity, due to its architecture, described in Section \ref{artslam}, and can be easily integrated and improved. Lastly, it is not bound to any framework, making it portable on different operative systems and customizable. ART-SLAM is available open source at \url{https://github.com/MatteoF94/ARTSLAM}.

\section{ART-SLAM} \label{artslam}

\subsection{System overview}

\begin{figure}[t]
      \centering
      \includegraphics[width=0.485\textwidth]{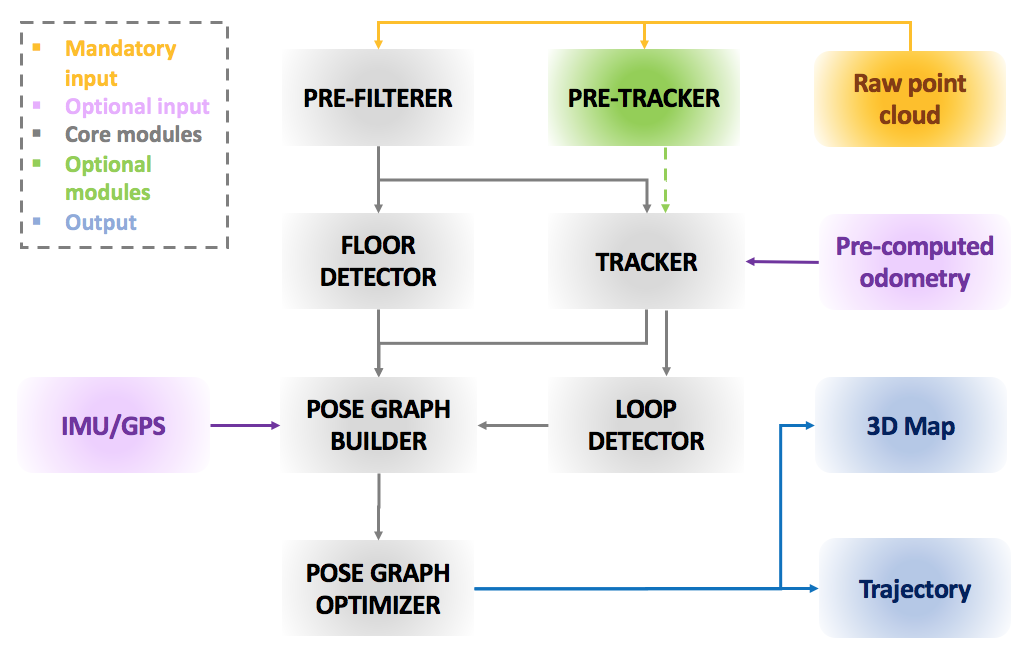}
      \caption{Architecture of the proposed system.}
      \label{system_architecture}
\end{figure}

An overview of the proposed framework is represented in Fig. \ref{system_architecture}. The system is composed by multiple distinct modules, which can be grouped into two main blocks. The first, mandatory (colored in gray), is the core of ART-SLAM, and it is formed by all the modules that perform SLAM on the input point clouds (orange, in figure). The other blocks of the proposed framework are optional, as they can be used to integrate the main system with data coming from different sensors or with re-processed inputs.

Given an incoming laser scan, the first step is to process it, in the \textit{pre-filterer}, to reduce its size and remove noisy points. The filtered cloud is then sent simultaneously to two modules. The most important one, the \textit{tracker}, estimates the current displacement of the robot by performing scan-to-scan matching with previous filtered scans. The other, \textit{floor detector}, finds the robot pose w.r.t. the ground, adding height and rotational consistency to the trajectory. The current pose estimate is sent, along with its corresponding point cloud, to the \textit{loop detector} module, which tries to find loops between new and previous point clouds, again performing scan-to-scan matching. Moreover, poses, loops and floor coefficients (estimated by the floor detector module) are used to build the pose graph, representing the trajectory of the robot. Lastly, the pose graph is optimized, to increase the poses accuracy.
 
\textit{IMU and GPS} data (pink in Fig. \ref{system_architecture}) can be integrated in the \textit{pose graph builder} module, to increase the accuracy of the estimated trajectory. Moreover, \textit{pre-computed odometry} (e.g., through a different sensor or system) can be fed to the tracker as initial guess for the scan matching. Lastly, a \textit{pre-tracker} module (green in Fig. \ref{system_architecture} performs multi-level scan-to-scan matching, to quickly estimate the motion of the robot before the tracking step: this estimate is broadcasted to the tracker, as initial guess of the scan-to-scan matching, to boost the accuracy and performance of the module. 

\subsection{System modularity}

\begin{figure}[t]
      \centering
      \includegraphics[width=0.48\textwidth]{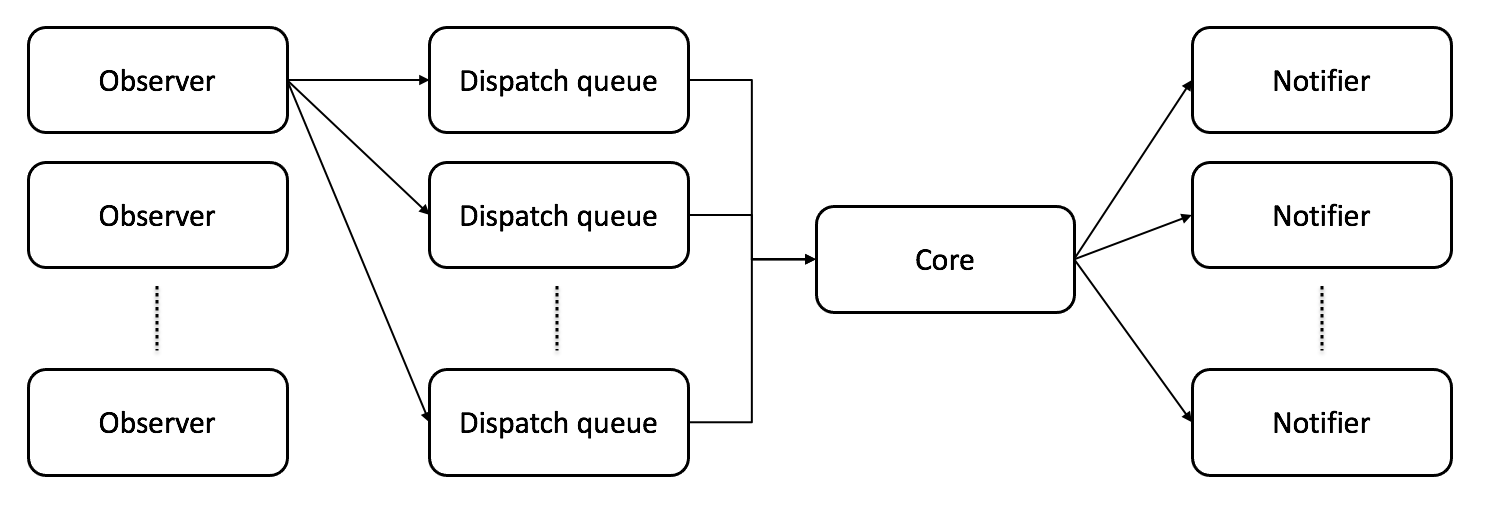}
      \caption{Inner structure of a single module. Observers capture data, dispatch queues store it, the core processes it and the notifiers broadcast the results.}
      \label{modularity}
\end{figure}

Differently to the majority of systems available in literature, our proposed framework is fully modular and its components work independently one from another. This is possible thanks to the \textit{register and dispatch} technique used to create the system. The structure of a module is represented in Fig. \ref{modularity}. It consists of one or multiple observers, one or multiple dispatch queues, one core, and one or multiple notifiers. Moreover, ART-SLAM is a zero-copy software, allowing the elaboration of large amount of data, while keeping it in memory.

Observers allow a module to capture data as soon as it is available, independently of the type. As data can arrive at a different rates than the time required for its processing, the observers put the received data into one or multiple dispatch queues, i.e., FIFO structures with the purpose of avoiding loss of incoming data. The core of the module is its main characteristic: it elaborates one datum per queue at time, extracting it from the relative dispatch queue. As soon as the core finishes its task, it gives the byproducts of the module to the notifier, which broadcasts them to all the modules in need. The advantage of using dispatch queues is the possibility of performing the same core task in parallel, on multiple threads, if it does not require temporal coherence.

This factory-like structure allows for an high degree of integration with auxiliary parts, improvements or third party modules. For example, if one would like to see the trajectory estimated by the tracker using its own implemented visualizer, it would only need to register the visualizer to the tracker and transform the broadcasted data in the desired way (data type conversion). Using this implementation brings multiple advantages over existing frameworks and middlewares, such as the possibility to execute in parallel independent core tasks, the portability (since it is embedded in ART-SLAM) and the high degree of customization.  

\subsection{Pre-filterer}
The pre-filterer module has the purpose of reducing the size of the input point cloud and to remove noise and outliers. Data reduction, or downsampling, is essential because, as stated in the introduction, scan matching on full point clouds can become computationally demanding if the size of the cloud is large enough (more than 20K points already proves to be a bottleneck on old devices). Downsampling can reduce point clouds by a factor 5, or even more, if needed, while retaining the spatial structure and density of the initial scan.

The clouds are then filtered, to remove outliers and noise points. This operation is more costly w.r.t. the downsampling task. To improve performances w.r.t. HDL \cite{koide2018portable}, we split the cloud into four pairs of octants and perform filtering on each separately, in parallel, obtaining a speedup of about $30\%$. After that, all the smaller clouds are combined together to form a larger, filtered point cloud, ready to be broadcasted to other modules.

\subsection{Tracker} \label{sec:tracker}
Short term data association, also known as pose tracking, establishes the motion between consecutive poses. The tracker adopts a keyframe-based approach to estimate the trajectory of the robot, performing scan-to-scan matching, using state-of-the-art algorithms (ICP \cite{besl1992method}, GICP \cite{segal2009generalized}, VGICP \cite{koide2020voxelized} and NDT \cite{biber2003normal}), depending on the user choice and the environment the robot is navigating.

\textit{Keyframes} are data structures describing the motion of robot in selected locations of its trajectory. They are described by multiple variables, used to collect data associated to the pose of the robot. 

In ART-SLAM, each keyframe contains a point cloud and the pose (odometry) estimated by the tracker, data which is also used for loop closure detection, pose graph construction and map creation. Other useful information contained in a keyframe are the timestamp associated to the point cloud, the estimated accumulated distance from the beginning of the trajectory and, if available, acceleration and orientation coming from other sensors, such as IMU.

To reduce the computational resources needed to perform SLAM, not all the filtered point clouds in input to the tracker become keyframes. With the exception of the first keyframe, which corresponds to the first point cloud received by the system, the other keyframes must satisfy at least one of the following criteria:
\begin{itemize}
    \item Be distant from the previous keyframe of a user-defined translation $\Delta trans$ meters
    \item Be rotated from the previous keyframe of a user-defined angle $\Delta orientation$ radians
    \item Have a difference in timestamps of a user-defined interval $\Delta T$ seconds
\end{itemize}
The thresholds $\Delta trans$, $\Delta orientation$ and $\Delta T$ are depend on the dataset and the type of trajectory to be estimated, and should be tuned accordingly to obtain a reasonable number of keyframes, as too few would decrease the accuracy of the SLAM system, and too many would decrease its performances. In indoor scenes, for example, $\Delta trans$ could be set to $0.2$ meters, while in large scale urban environments $\Delta trans > 2$ meters. 

Given the cloud corresponding to the current keyframe $K_{n}$ and the available new filtered point cloud in input $c_{t}$, scan-to-scan matching is performed between them, to find their relative motion. The algorithm requires an initial guess of the motion, to boost performance and accuracy. There are two choices for the initial guess: either this is available through other means (e.g. from odometry estimated using the pretracker module), or a constant motion model is assumed, and the previous relative transformation is used (the one computed between the point cloud of the current keyframe $K_{n}$ and the previous filtered point cloud $c_{t-1}$). 

Usually, algorithms for point cloud-based tracking find the relative motion between consecutive clouds, $c_{t-1}$ and $c_{t}$, and then compose this transformation with the previous ones. This method may seem more accurate, but it accumulates error the more distant the clouds are from the current keyframe. In ART-SLAM, instead, the motion of the robot is always referred w.r.t. the keyframe closest in time, and the previous motion is taken into consideration only when estimating the guess for scan matching, as in the previous paragraph. 

This approach, which is unique to ART-SLAM (HDL does not have it) also enables the system to skip input clouds (meaning that no scan matching is performed) if pre-computed odometry is available. The latter, even if not completely accurate, allows to immediately check if the current point cloud is a candidate for the selection of a new keyframe. If it is not, the tracker does not perform scan matching, and the relative transformation between the current keyframe and the pre-computed odometry is saved to be used as motion guess in the next iteration. Skipping the scan matching step greatly benefits the performance of the tracker, as it allows to obtain accurate results in real-time.

Once the tracker has detected a point cloud which satisfies the keyframe creation criteria described above, a new keyframe is built and it is broadcasted to the loop closure detection and pose-graph builder modules.

\subsection{Pre-tracker} \label{sec:pretracker}
Although temporally the pre-tracker module works in parallel with the pre-filterer, to understand its working principle we described first the pre-filterer and the tracker modules.

Indeed, performing alignment between two full scale clouds would result in the best transformation estimate, as all the 3D points are accounted for. However, this approach is often unsuitable for real-time application, especially on low-end devices. Scan matching can be aided, as described in Subsection \ref{sec:tracker}, with an initial guess, already available, which can lower the time needed to reach matching convergence and increase the accuracy of the estimated transformation between two point clouds.

To compute a viable initial guess to give the tracker, we created a pre-tracker module, which performs multi-scale scan matching, working as follows. First, the same point cloud given in input to the pre-filterer is fed to the pre-tracker. Here, it is heavily downsampled, reducing it to less than 10\% the number of its initial elements. The reduced point cloud is then used to perform scan-to-scan matching with a previously downsampled cloud. This alignment is fast, due to the reduced size of the point clouds, even if not as accurate as if it was done with non-downsampled clouds. 

If the number of elements of the initial point cloud is relatively large (greater than 60K 3D points), the transformation obtained through the step above is immediately broadcasted to the tracker, to be used as initial guess in the current scan matching. On the other hand, if the starting cloud size is relatively small, the whole procedure can be repeated with a different scale. The point cloud is downsampled with a scale factor lower than the one used in the first phase of the pre-tracker, to obtain a reduced cloud, with size greater than the one obtained through the first phase. Again, the obtained point cloud is used to perform scan-to-scan matching with a previously downsampled cloud, in order to obtain a fast result, but more accurate than the transformation obtained in the first phase. At this point, the relative motion resulting from the second phase is broadcasted to the tracker, to be used, as already stated, as initial guess in the current scan matching.

The adoption of a pre-tracker proves to be beneficial not only in terms of accuracy, as it gives the tracker an initial guess close to the true transformation, but also in terms of performance, as it allows the tracker to skip some frames, which would not be unused anyway in the SLAM system, as described in Subsection \ref{sec:tracker}.

\subsection{Floor detection}
To enforce height and orientation consistency in the trajectory, filtered point clouds are processed to find the ground plane in them. This can be modeled as a four dimensional vector $GP(a,b,c,d)$ represented by the plane equation $a*x + b*y + c*z + d = 0$. 

Floor detection handles multiple scenarios, such as planar or planar-like motion (e.g., urban road), rough terrains (e.g. rocky paths) and environments with ascents and descents. While HDL \cite{koide2018portable} deals only with the planar motion, in ART-SLAM all the scenarios are considered.

In the first case, planar motion, the floor detector module takes a point cloud and manipulates it in the following way. As, intuitively, the ground can be found within a small region of the input scan, the first step performed by the floor detector is clipping the cloud within an acceptable range of search. This step greatly reduces the cloud size, boosting performances when searching for the floor. Then, the clipped output is filtered to eliminate points whose normal is highly non-vertical. This is done to avoid possible mistakes due to planar-like surfaces in the environment, such as walls and buildings. Lastly, Random Sample Consensus (RANSAC) for plane detection is done on the filtered laser scan, to detect and estimate the ground plane coefficients. 

When dealing with rough terrains, a floor cannot be found in the previous way, as no planar structures can be detected with RANSAC. The input scan is further clipped, this time not vertically but horizontally: only the 3D points within a threshold distance from the center of the cloud are kept. This is done to trim the cloud to be as close as possible to the robot, to remove outlier objects such as big rocks, logs, or anything which is not planar-like. The few remaining points are then used to perform closed form plane fitting with the least squares method. 

Once, and if, the parameters $\{a,b,c,d\}$ are found, they are broadcasted, together with the timestamp associated to the corresponding point cloud, to the pose graph builder module.

The last scenario is trickier to identify just by using a point cloud, as inclined planes are parallel to the robot wheels and cannot be distinguished from non-inclined planes by just using point clouds for detection. The process of discovery of inclined planes takes place in the pose-graph builder module. When a set of floor coefficient $\{a,b,c,d\}$ is associated to a keyframe, the builder checks if there is a noticeable change (user-defined) in vertical orientation w.r.t. the previous keyframe. If affirmative, it means that there has been a change in slope in the trajectory of the robot, and an inclined ground plane has been detected.

\subsection{Loop closure}
While moving, the robot may return to a place which was previously visited, forming a loop in its trajectory. Finding loops adds motion constraints in the estimated robot poses, correcting drifts and estimation errors. The hard part about loop closure is not asserting the presence of a loop, which can be accomplished via simple scan matching, but detecting when loop closure is even a possibility. To do this we need to decide when and where to look. In ART-SLAM, detection is performed in three consecutive steps, to efficiently search for loops within the collected keyframes.

First, each time a keyframe $K_{query}$ is available, it is compared against all the previous existing keyframes, creating a query and candidates problem structure. Instead of performing scan-to-scan matching between all the possible pairs $\{K_{new}, K_{candidate}\}$, an odometry based selection is performed. If $K_{query}$ and $K_{candidate}$ are too close in terms of trajectory, meaning that they have a low accumulated distance, they cannot be considered candidates, as it is unlikely that two keyframes, corresponding to point clouds acquired shortly one from the other, would result in a loop. Moreover, the loop detector checks if the position, estimated through tracking, of $K_{candidate}$ is in the neighborhood of the pose corresponding $K_{query}$, within a threshold range, which accounts for drift errors induced by the tracker module. If $K_{query}$ and $K_{candidate}$ satisfy these constraints, meaning that they are sufficiently close in space and far in time, they can be considered a loop closure candidate pair, to be fed to the next phase. 

Once all the candidate pairs have been found, they must be further thinned down, to avoid unnecessary computation and wasted resources. The approach proposed in \cite{gkim-2018-iros} converts point clouds in 2D polar grids, and efficiently compares them using a KD-tree to select the $k$ most similar ones to a given point cloud query. The second phase for efficient loop closure detection in ART-SLAM adopts this method, by comparing the 2D polar grid of the point cloud associated to the query keyframe with the 2D polar grids corresponding to the candidate keyframes. At the end of this step, only $k$ candidate pairs for loop closure remains, ready to be used in the next, last step.

The few number of candidates allows for scan-to-scan matching on each pair of point clouds, to obtain a set of relative motions. All the transformations are then compared to find the best one, i.e., the one computed with highest accuracy, and corresponding to the smallest distance between all the pairs $K_{query}$ and $K_{candidate}$. If a best match is found, a new loop has been detected, and it is added to the pose graph as a new constraint.

Differently from HDL \cite{koide2018portable}, where only the first and last steps are performed, in ART-SLAM, the addition of the Scan Context method allows for scalable and efficient loop closure. Indeed, as the length of the trajectory to be estimated increases, the number of pairs to be checked for loop closures also grows in size, because more and more keyframes are added. However, the first two steps are very fast operations, with the former consisting mainly in a matrix multiplication and the latter being proved to be scalable in \cite{gkim-2018-iros}. Moreover, the 2D polar grids are pre-computed when inserting the keyframes in the pose graph, further decreasing the computation time needed by the loop closure module. At the end of the second phase, there will always be at most $k$ candidate pairs, independently from the number of keyframes to check, making this three-phased approach suitable for efficient loop closure detection.

\subsection{Pose graph building and optimization}
As mentioned in the introduction and in the description of the system, our framework is a form of graph SLAM \cite{grisetti2010tutorial}. In graph SLAM, the poses of the robot are modeled as nodes in a graph, named \textit{pose graph}, and labeled with their position in the environment. The nodes are connected with edges representing spatial constraints between poses, resulting from sensor measurements (e.g. IMU or GPS) or scene elements, as the floor coefficients in our case. 

Each node in the pose graph represents a robot position and a measurement (the point cloud) acquired at that position; moreover, each node is associated to the corresponding keyframe. An edge between two nodes consists in a probability distribution over the relative transformation of the robot poses corresponding to the nodes. These transformations are either odometry measurements given by the tracker module, between sequential positions, or are determined by aligning the sensor measurements acquired between two keyframes. Because of the noise corrupting the sensors and the drift in the robot odometry, the associated edges represent soft constraints and are not fixed. It is, however, possible to insert absolute constraints, which cannot be modified in any way. Examples of these constraints are floor coefficients, GPS or IMU data, although they can also be set, instead, as non-absolute constraints, to account for the uncertainty of the sensors or the measurement. Moreover, edges can be added when performing loop detection and closure, between non-consecutive nodes in the graph. 

\begin{table}[t]
\vspace*{10pt}
\caption{Parameters used for the experimental validation, for reproducibility, as described in \cite{amigoni2009insightful}.}
\label{params}
\begin{center}
\begin{tabular}{|c||l||c|}
\hline
Module & Parameter name & Value\\
\hline
\multirow{4}{*}{\textit{pre-filterer}} 
& Downsample method & VOXELGRID \\
& Downsample resolution & 0.25 [m] \\
& Outlier removal method & RADIUS \\
& Radius & 0.4 [m]\\
\hline
\multirow{3}{*}{\textit{tracker}} 
& $\Delta$Trans keyframes & 5.0 [m] \\
& $\Delta$Angle keyframes & 0.25 [rad] \\
& $\Delta$Time keyframes & 1.0 [sec] \\
\hline
\multirow{2}{*}{\textit{loop detector}} 
& Loop closure search radius  & 40.0 [m] \\
& Loop closure min. distance & 25.0 [m] \\
\hline
\multirow{3}{*}{\textit{scan matching}} 
& Registration method & FAST\_GICP \\
& Max. iterations & 64 \\
& Transformation epsilon & 0.1 \\
\hline
\end{tabular}
\end{center}
\end{table}

The structure of the pose graph is given to optimization algorithms to compute the optimal trajectory which satisfies all the sensors and motion constraints, giving high accuracy estimates, while elaborating a large number of poses. In our implementation, we use the g2o optimization framework \cite{grisetti2011g2o}, as it proves to be fast and accurate over long trajectories. Moreover g2o allows for the insertion of custom elements in the pose graph, and as such, it is an optimal solution for our modular system.

The choice of modeling the trajectory of the robot as a graph is motivated by multiple reasons. First, the poses can be easily stored and visualized at any time, including also the additional information contained in the graph (edges, special nodes created from data of other sensors or ground). Furthermore, the ability of the g2o \cite{grisetti2011g2o} graph optimizer to calculate the optimal minimum cost function, to satisfy all the constraints, gives our system a high accuracy and consistency in the estimation technique for solving the SLAM problem. This is important in the case of autonomous driving, particularly in dealing with large scale environments. More details about the advantages of using graphs are in \cite{grisetti2010tutorial}.

\section{Experimental validation of the system} \label{experiments}
\subsection{Setup}
The proposed system is compared with other methods for point cloud-based SLAM: LOAM \cite{zhang2014loam}, LeGO-LOAM \cite{shan2018lego}, A-LOAM, LeGO-LOAM-BOR and HDL \cite{koide2018portable}, with A-LOAM and LeGO-LOAM-BOR being two advanced versions of LeGO-LOAM (code improvement and re-engineering of LeGO-LOAM). We evaluate these systems in four scenarios: three coming from the KITTI dataset \cite{Geiger2012CVPR} \cite{Geiger2013IJRR}, corresponding to a short, a medium and a long sequences, respectively, and one from the RADIATE dataset \cite{sheeny2020radiate}, representing a medium sequence with no loops.

\begin{figure}[t]
      \vspace*{10pt}
      \hspace{0.17em}
      \includegraphics[width=0.4\textwidth]{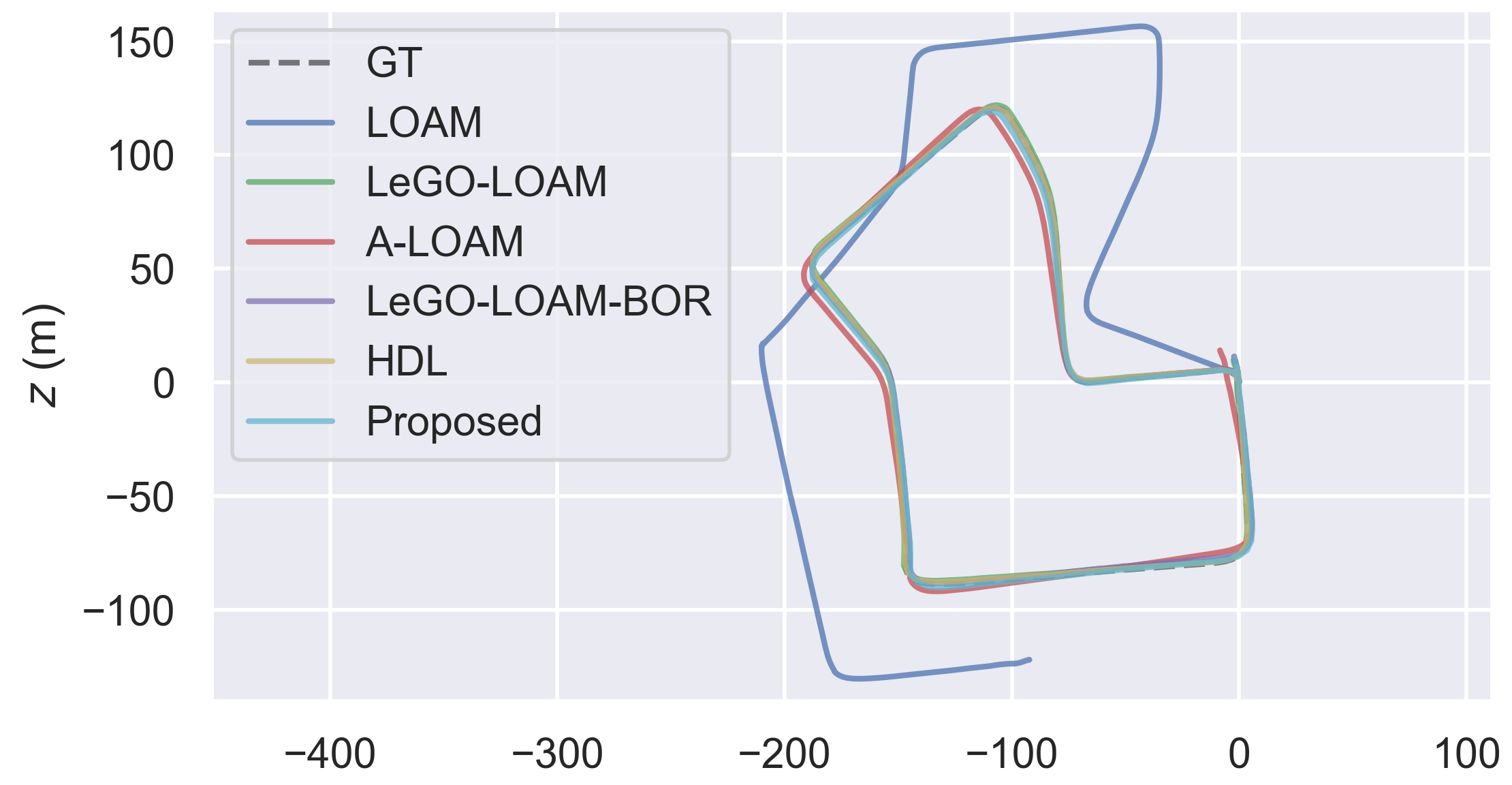}
      \caption{Comparison between the trajectories estimated by LOAM \cite{zhang2014loam}, LeGO-LOAM \cite{shan2018lego}, A-LOAM, LeGO-LOAM-BOR, HDL \cite{koide2018portable} and the proposed system, on Sequence 07 of the KITTI odometry dataset \cite{Geiger2012CVPR}.}
      \label{07_comp}
\end{figure}

\begin{table}[t]
\vspace*{10pt}
\caption{ATE on Sequence 07 of KITTI odometry dataset \cite{Geiger2012CVPR}.}
\label{07_table}
\begin{center}
\begin{tabular}{|c||c||c||c|}
\hline
ATE[m] & MEAN & RMSE & STD\\
\hline
\textit{LOAM} & $>$10 & $>$10 & $>$10\\
\hline
\textit{LeGO-LOAM} & 1.191 & 1.309 & 0.546\\
\hline
\textit{A-LOAM} & 2.467 & 2.741 & 1.195\\
\hline
\textit{LeGO-LOAM-BOR} & 1.604 & 1.807 & 0.832\\
\hline
\textit{HDL} & 0.407 & 0.439 & \textbf{0.145}\\
\hline
\textit{ART-SLAM (proposed)} & \textbf{0.405} & \textbf{0.435} & 0.157\\
\hline
\end{tabular}
\end{center}
\end{table}

LOAM, LeGO-LOAM, A-LOAM and LeGO-LOAM-BOR do not require particular parameter tuning, although they need a custom implementation of one of their modules, (point cloud projection), depending on the laser sensor used. In our tests, we changed such parameters accordingly to the sensor corresponding to the used datasets, as suggested by the authors of LeGO-LOAM. On the other hand, HDL and ART-SLAM share the same configuration parameters, e.g., keyframe selection thresholds and pre-filtering methods. Table \ref{params} shows the most important parameters used in the experiments, for both HDL and ART-SLAM, to allow reproducibility. As few systems only work on point clouds, for fair comparison, we perform SLAM only using point clouds, without exploiting data coming from other sensors, such as IMU or GPS. 

Experiments are tested on a 2012 Dell 64-bit laptop with Intel(R) Core(TM) i5-3337U CPU @ 1.80GHz x 4 cores, each with 3072Kb of cache size.

\subsection{Comparison and results}

To evaluate the systems we compute the absolute trajectory error (ATE). This metric measures the difference between points of the true and the estimated trajectory. As a pre-processing step, we associate the estimated poses with ground truth poses using the timestamps and point cloud indices. We also include a visual evaluation of the estimated trajectory and show the reconstructed 3D map in two out of the four considered scenarios. Nevertheless, all the systems run in real-time, meaning that they can process data at its acquisition rate, with the exception of HDL, which requires more time when performing both tracking and loop closure detection (about two to three times the acquisition rate).

\begin{figure}[t]
      \vspace*{10pt}
      \hspace{0.15em}
      \includegraphics[width=0.40\textwidth]{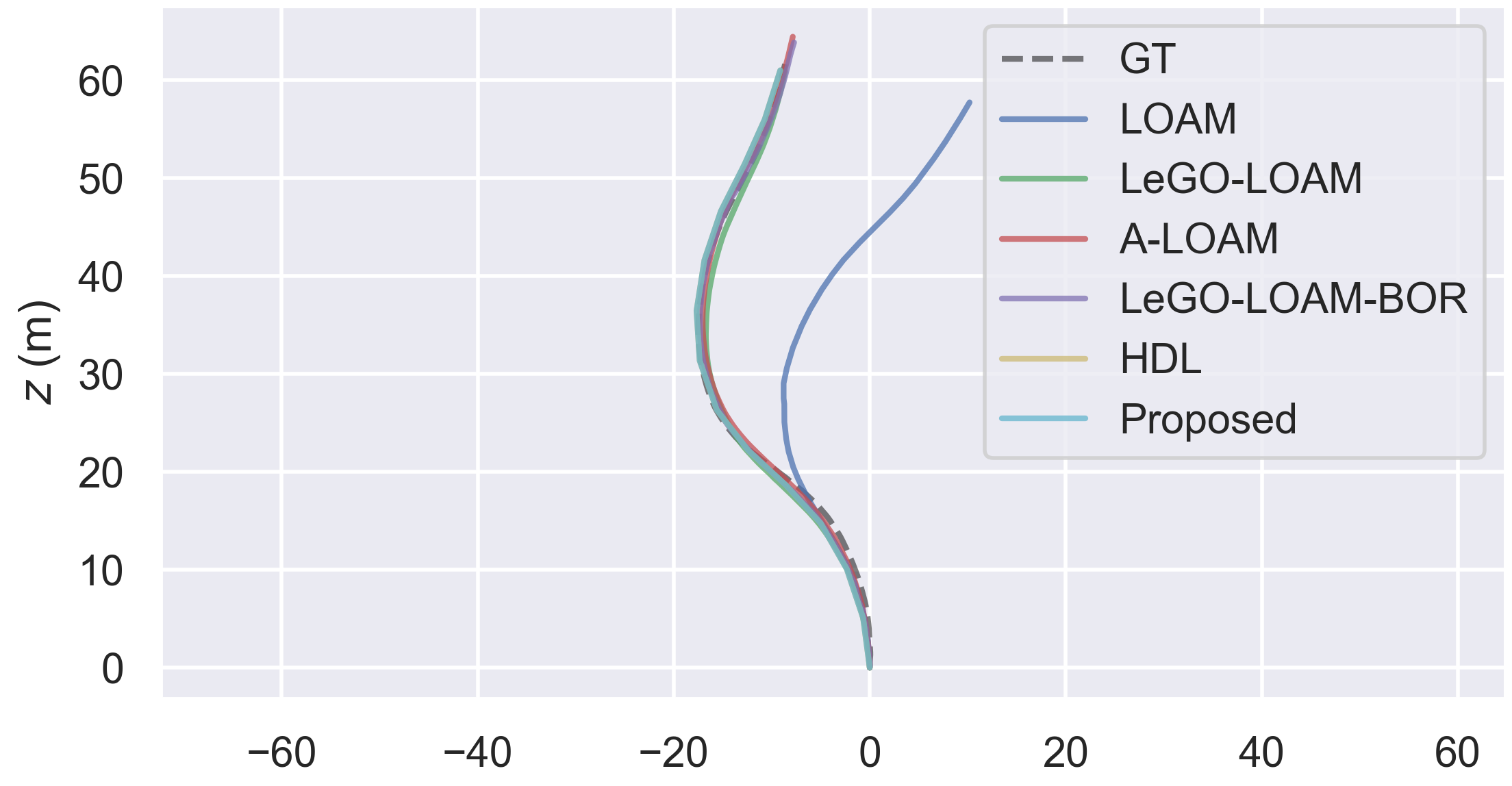}
      \caption{Comparison between the trajectories estimated by LOAM \cite{zhang2014loam}, LeGO-LOAM \cite{shan2018lego}, A-LOAM, LeGO-LOAM-BOR, HDL \cite{koide2018portable} and the proposed system, on city Sequence 05 of the KITTI raw dataset \cite{Geiger2013IJRR}.}
      \label{26_comp}
\end{figure}

\begin{table}[t]
\vspace*{10pt}
\caption{ATE on the short city Sequence 05 of KITTI raw dataset \cite{Geiger2013IJRR}.}
\label{26_table}
\begin{center}
\begin{tabular}{|c||c||c||c|}
\hline
ATE[m] & MEAN & RMSE & STD\\
\hline
\textit{LOAM} & $>$5 & $>$5 & $>$5\\
\hline
\textit{LeGO-LOAM} & 0.707 & 0.768 & 0.300\\
\hline
\textit{A-LOAM} & 0.938 & 1.044 & 0.459\\
\hline
\textit{LeGO-LOAM-BOR} & 1.094 & 1.169 & 0.409\\
\hline
\textit{HDL} & \textbf{0.703} & 0.798 & 0.376\\
\hline
\textit{ART-SLAM (proposed)} & \textbf{0.703} & \textbf{0.760} & \textbf{0.330}\\
\hline
\end{tabular}
\end{center}
\end{table}

Fig. \ref{07_comp} shows the estimated trajectories on Sequence 07 of the KITTI odometry dataset \cite{Geiger2012CVPR}. All the methods considered for comparison, with the exception of LOAM, accurately follow the ground truth trajectory, correctly finding the loop and optimizing the poses. LOAM, instead, quickly drifts apart from the true trajectory: this is caused by the fact that no loop closure is performed, differently from the other systems. Table \ref{07_table} further details the obtained results, as it represents the mean, root mean squared error (RMSE) and standard deviation (STD) of the absolute trajectory error, in meters. The proposed system has highest accuracy, along with HDL. It should not come as a surprise, since those methods rely on full point cloud scan-to-scan matching, while the other methods rely on tracking and matching 3D features extracted from point clouds. 

After having dealt with a large sequence with the presence of loops, we also evaluated the systems on a shorter sequence. As short datasets do not have a ground truth, we use, instead, GPS data, provided along with the point clouds. Fig. \ref{26_comp} shows the estimated trajectories on city Sequence 05 of the KITTI raw dataset \cite{Geiger2013IJRR}. As before, all the methods except for LOAM, accurately represent the ground truth, with small errors in the trajectory. It should not come as a surprise that the results are more or less the same, as for short trajectories tracking is performed a limited amount of times, and there is not enough distance to suffer from accumulated errors. Table \ref{26_table} shows the ATE statistics, in meters. As before, all the systems, with the exception of LOAM, show good results, accurately following the GPS signal, here used as ground truth due to its relatively high accuracy.

\begin{figure}[t]
      \vspace*{10pt}
      \hspace{0.15em}
      \includegraphics[width=0.4\textwidth]{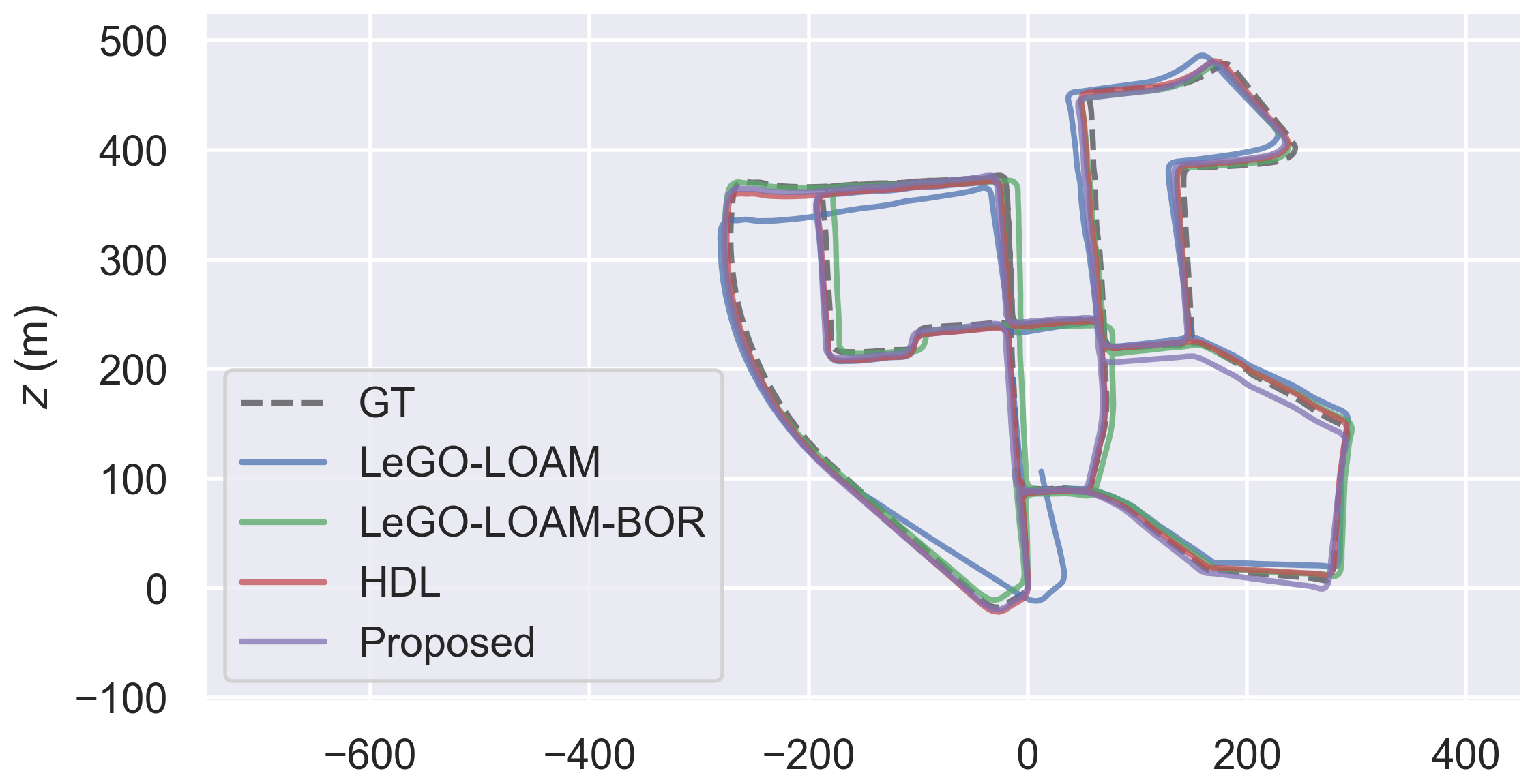}
      \caption{Comparison between the trajectories estimated by LeGO-LOAM \cite{shan2018lego}, LOAM \cite{zhang2014loam}, HDL \cite{koide2018portable} and the proposed system, on sequence 00 of the KITTI odometry dataset \cite{Geiger2012CVPR}.}
      \label{00_comp}
\end{figure}

\begin{table}[t]
\vspace*{10pt}
\caption{ATE on Sequence 00 of KITTI odometry dataset \cite{Geiger2012CVPR}.}
\label{00_table}
\begin{center}
\begin{tabular}{|c||c||c||c|}
\hline
ATE[m] & MEAN & RMSE & STD\\
\hline
\textit{LeGO-LOAM} & 9.537 & 11.666 & 6.718\\
\hline
\textit{LeGO-LOAM-BOR} & 6.240 & 6.613 & 2.188\\
\hline
\textit{HDL} & \textbf{1.078} & \textbf{1.224} & 0.579\\
\hline
\textit{ART-SLAM (proposed)} & 1.119 & 1.352 & \textbf{0.539}\\
\hline
\end{tabular}
\end{center}
\end{table}

\begin{figure}[t]
      \centering
      \vspace*{10pt}
      \includegraphics[width=0.25\textwidth]{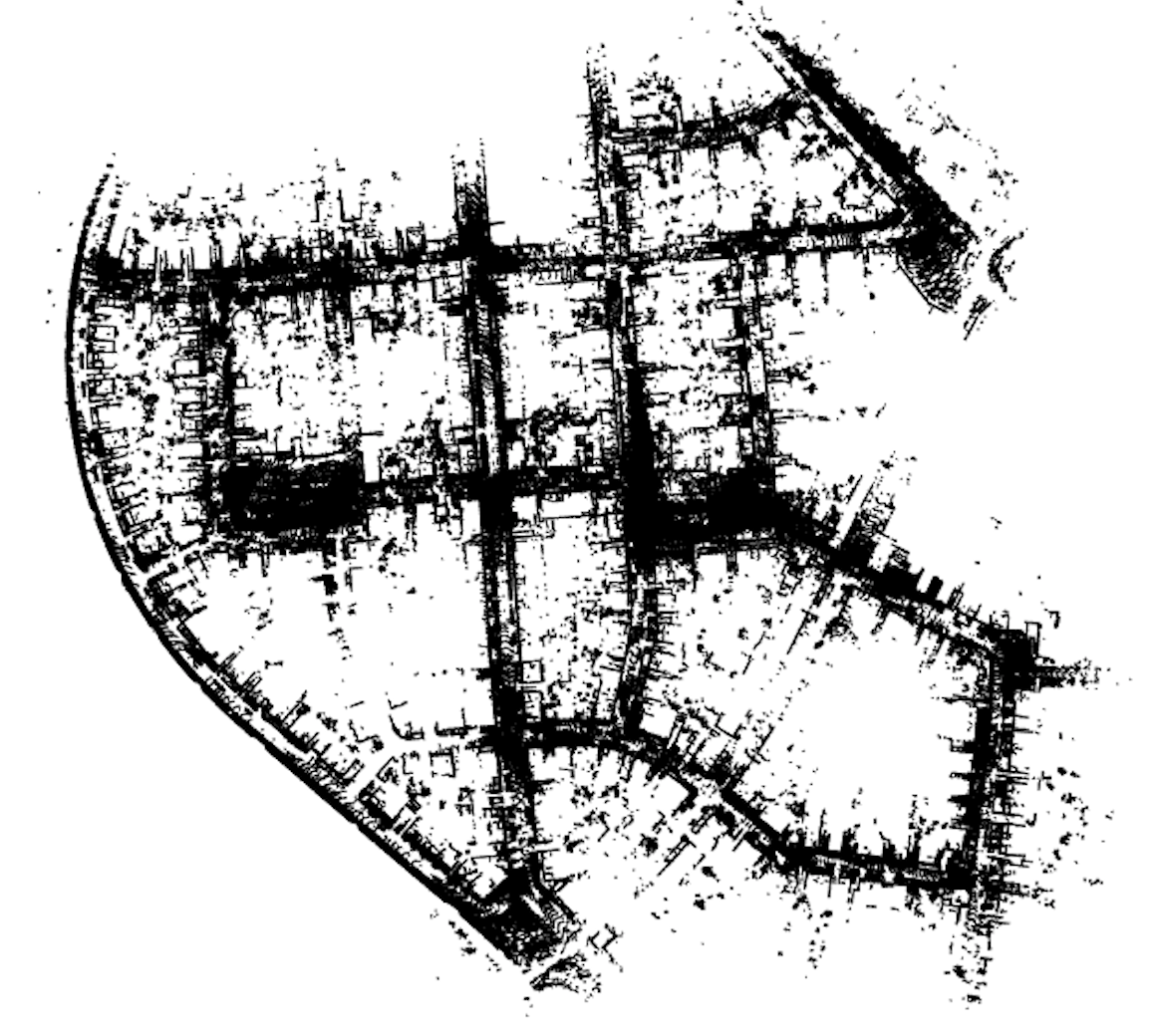}
      \caption{Map corresponding to Sequence 00 of the KITTI odometry dataset \cite{Geiger2012CVPR}, built by the proposed algorithm.}
      \label{00_map}
\end{figure}

\begin{figure}[t]
      \centering
      \vspace*{10pt}
      \includegraphics[width=0.28\textwidth]{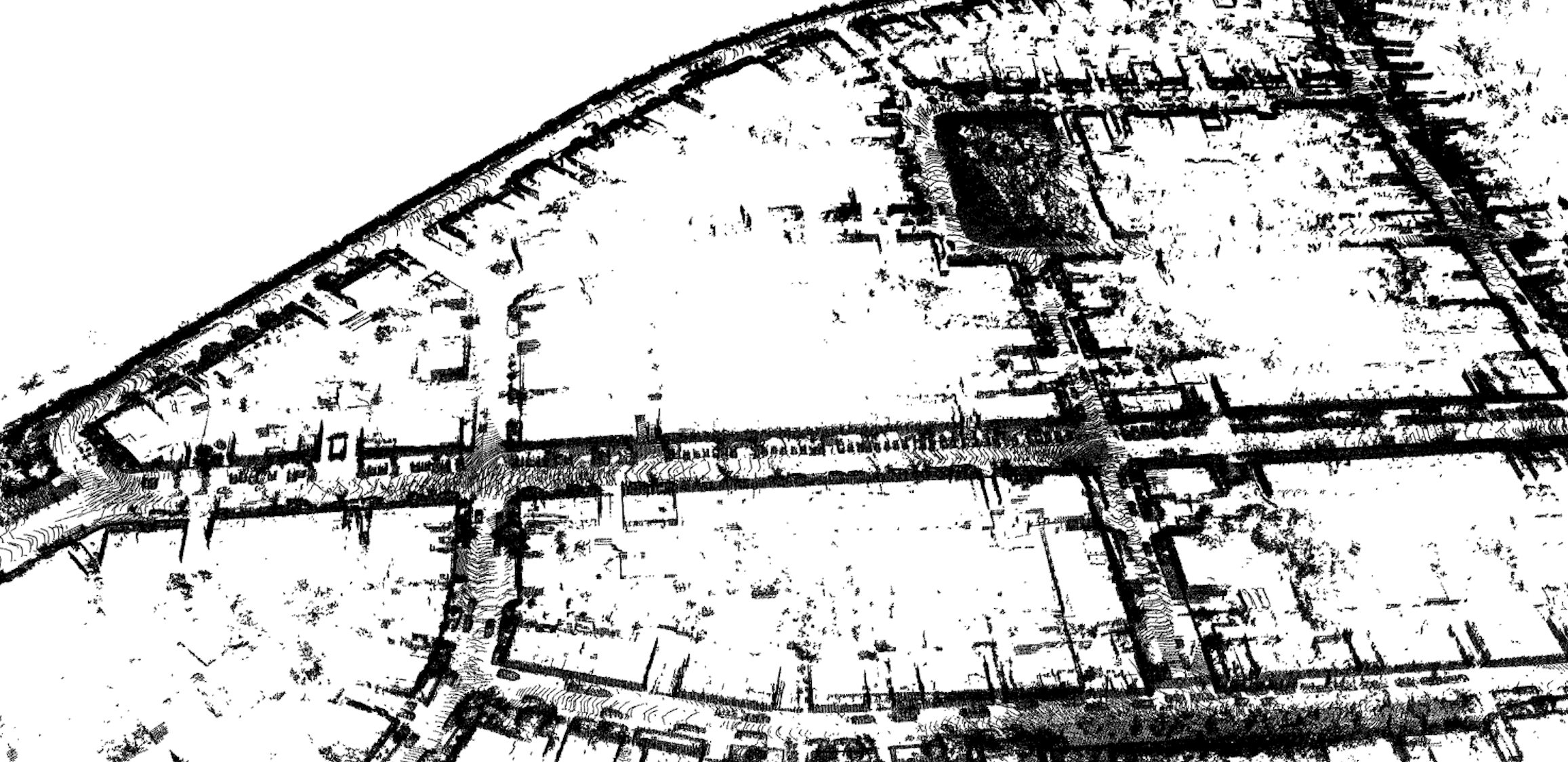}
      \caption{Detailed area of the map corresponding to Sequence 00 of the KITTI odometry dataset \cite{Geiger2012CVPR}, built by the proposed algorithm.}
      \label{00_detail}
\end{figure}

In Fig. \ref{00_comp}, we show the behavior of ART-SLAM on one of most complex sequences, i.e, Sequence 00 of the KITTI odometry dataset. In the comparison, we did not include LOAM and A-LOAM, which estimated trajectories are far off the ground truth, and would have cluttered the figure. Fig. \ref{00_map} and Fig. \ref{00_detail} show the map reconstructed by ART-SLAM and a detailed area of it, respectively. From Table \ref{00_table}, one can see the high degree of accuracy achieved by the proposed system, reaching low translation error almost on par with HDL.

\begin{figure}[t]
      \centering
      \vspace*{10pt}
      \includegraphics[width=0.25\textwidth]{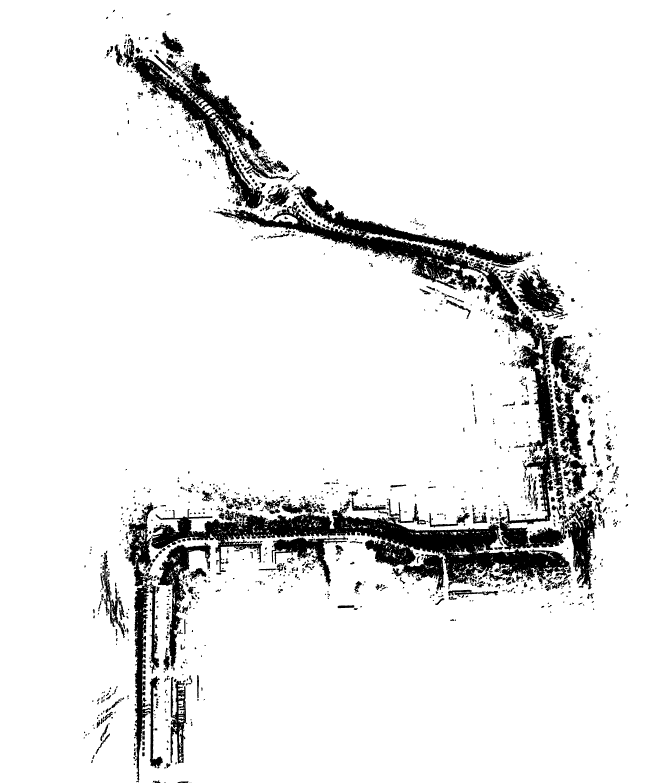}
      \caption{Map corresponding to City Sequence 01 of the RADIATE dataset \cite{sheeny2020radiate}, built by the proposed algorithm.}
      \label{c01_map}
\end{figure}

\begin{figure}[t]
      \centering
      \includegraphics[width=0.25\textwidth]{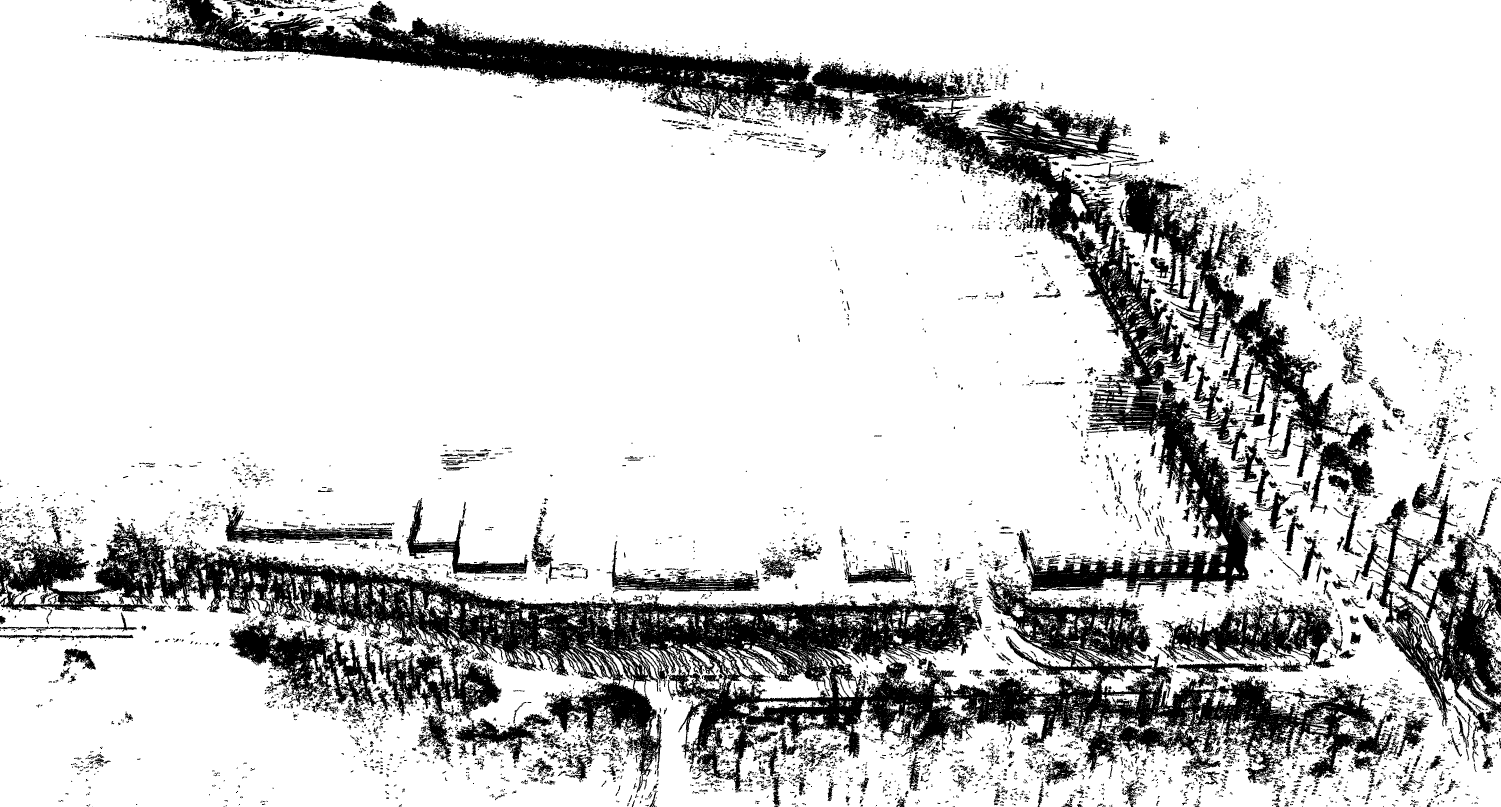}
      \caption{Detailed area of the map corresponding to City Sequence 01 of the RADIATE dataset \cite{sheeny2020radiate}, built by the proposed algorithm.}
      \label{c01_detail}
\end{figure}

\begin{figure}[t]
      \centering
      \vspace*{10pt}
      \includegraphics[width=0.48\textwidth]{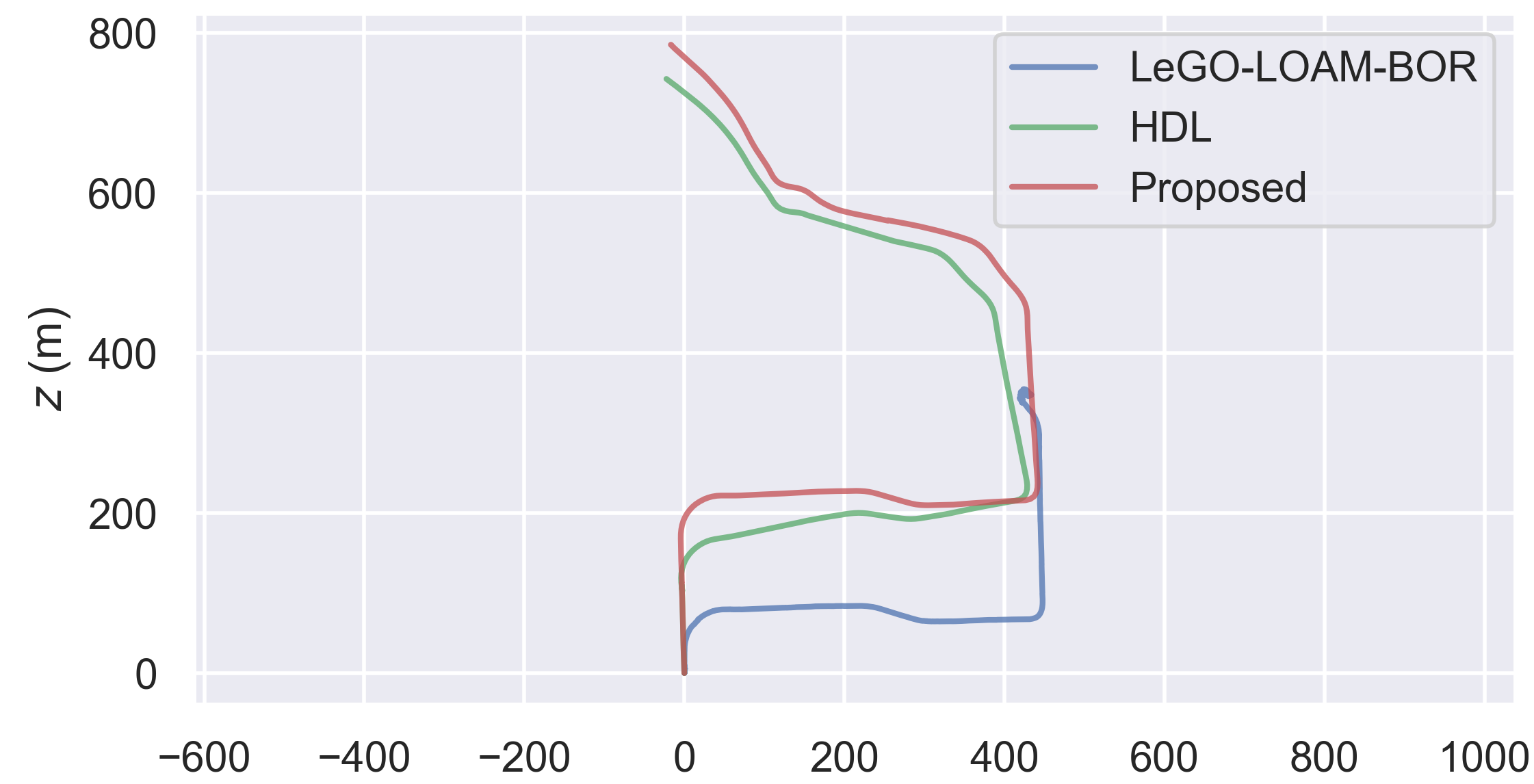}
      \caption{Comparison between the trajectories estimated by LeGO-LOAM-BOR, HDL \cite{koide2018portable} and the proposed system, on City Sequence 01 of the RADIATE dataset \cite{sheeny2020radiate}.}
      \label{c01_comp}
\end{figure}

Lastly, we show a visual evaluation of the accuracy achieved by ART-SLAM on the City 01 Sequence of the RADIATE dataset \cite{sheeny2020radiate}. This sequence is relatively long and does not contain loop closures, increasing the difficulty of estimating the robot trajectory. As there is no ground truth, we perform a visual inspection of the created 3D map, to check for inconsistencies. The map obtained through ART-SLAM, completely visible in Fig. \ref{c01_map} and detailed in Fig. \ref{c01_detail}, shows a noticeable coherence with the structure of the road and the elements within it, once again proving the accuracy of the proposed system. Moreover, Fig. \ref{c01_comp} shows a comparison of the obtained trajectory with HDL and LeGO-LOAM-BOR (the other systems could not run the scenario), proving that ART-SLAM is accurate even when no loop closure is available.

\section{Conclusions} \label{conclusions}

We have proposed ART-SLAM, a fast and ground-optimized LiDAR odometry and mapping method, able to perform pose estimation of moving robots in complex environments. ART-SLAM, differently from state-of-the-art systems, is not bound to any framework, and can be easily ported on any device. It is also efficiently improvable and extendable, due the independent nature of its modules, and it includes many upgrades w.r.t. existing similar systems, such as pre-tracking, smart loop closure and optimized loop detection. The proposed method is evaluated on a series of datasets corresponding to outdoor environments, representing either short, medium or long sequences. The results show that ART-SLAM can achieve similar or better accuracy when compared with the state-of-the-art, with reduced computational cost w.r.t. high accuracy systems. The proposed system proves to be as fast as feature-based systems, meaning real-time or near real-time performance, and accurate as full point clouds scan matching methods.

\addtolength{\textheight}{-0cm}   





\printbibliography

\end{document}